\documentclass[runningheads]{llncs}

\usepackage{eccv}

\usepackage{eccvabbrv}
\usepackage{pifont}
\usepackage{graphicx}
\usepackage{booktabs}
\usepackage{multirow}
\usepackage{makecell}
\usepackage{bbding}
\usepackage{hyperref}
\usepackage[accsupp]{axessibility}  

\usepackage{hyperref}

\usepackage{orcidlink}
\usepackage[capitalize]{cleveref}
\crefname{figure}{fig}{figures}
\Crefname{figure}{Fig}{Figures}

\usepackage{floatrow}
\newfloatcommand{capbtabbox}{table}[][\FBwidth]

\hypersetup{hidelinks,
	colorlinks=true,
	allcolors=red,
	pdfstartview=Fit,
	breaklinks=true}

\begin{document}

\title{Category-level Object Detection, Pose Estimation and Reconstruction from Stereo Images} 

\titlerunning{Coders}

\author{Chuanrui Zhang\inst{1,2\ast} \and
Yonggen Ling \inst{2\ast \dag} \and
Minglei Lu \inst{2} \and
Minghan Qin \inst{1} \and
Haoqian Wang \inst{1 \dag}}

\authorrunning{C.~Zhang et al.}

\institute{Tsinghua University, Beijing, China \and Tencent Robotics X, Shenzhen, China \\
\email{zhang-cr22@mails.tsinghua.edu.cn}  \\
\email{\{rolandling, mingleilu\}@tencent.com} \\
\email{qmh21@mails.tsinghua.edu.cn}  \\
\email{wanghaoqian@tsinghua.edu.cn}}


\maketitle

{\let\thefootnote\relax\footnotetext{{$^{\ast}$ Equal contribution.
Work done while Chuanrui Zhang is an intern at Tencent Robotics X.\\ 
$\dag$ Corresponding authors.}}}

\begin{abstract}
    We study the 3D object understanding task for manipulating everyday objects with different material properties (diffuse, specular, transparent and mixed).
    Existing monocular and RGB-D methods suffer from scale ambiguity due to missing or imprecise depth measurements.
    We present \textbf{CODERS}, a one-stage approach for \textbf{C}ategory-level \textbf{O}bject \textbf{D}etection, pose \textbf{E}stimation and \textbf{R}econstruction from \textbf{S}tereo images. 
    The base of our pipeline is an Implicit Stereo Matching module that combines stereo image features with 3D position information.
    Concatenating this presented module and the following transform-decoder architecture leads to end-to-end learning of multiple tasks required by robot manipulation.
    Our approach significantly outperforms all competing methods in the public TOD dataset. 
    Furthermore, trained on simulated data, CODERS generalize well to unseen category-level object instances in real-world robot manipulation experiments.
    Our dataset, code, and demos will be available at \href{https://xingyoujun.github.io/coders}{https://xingyoujun.github.io/coders}.
\end{abstract}

\section{Introduction}
\label{sec:intro}
\begin{figure}[tb]
  \centering
  \includegraphics[width=1.0\textwidth]{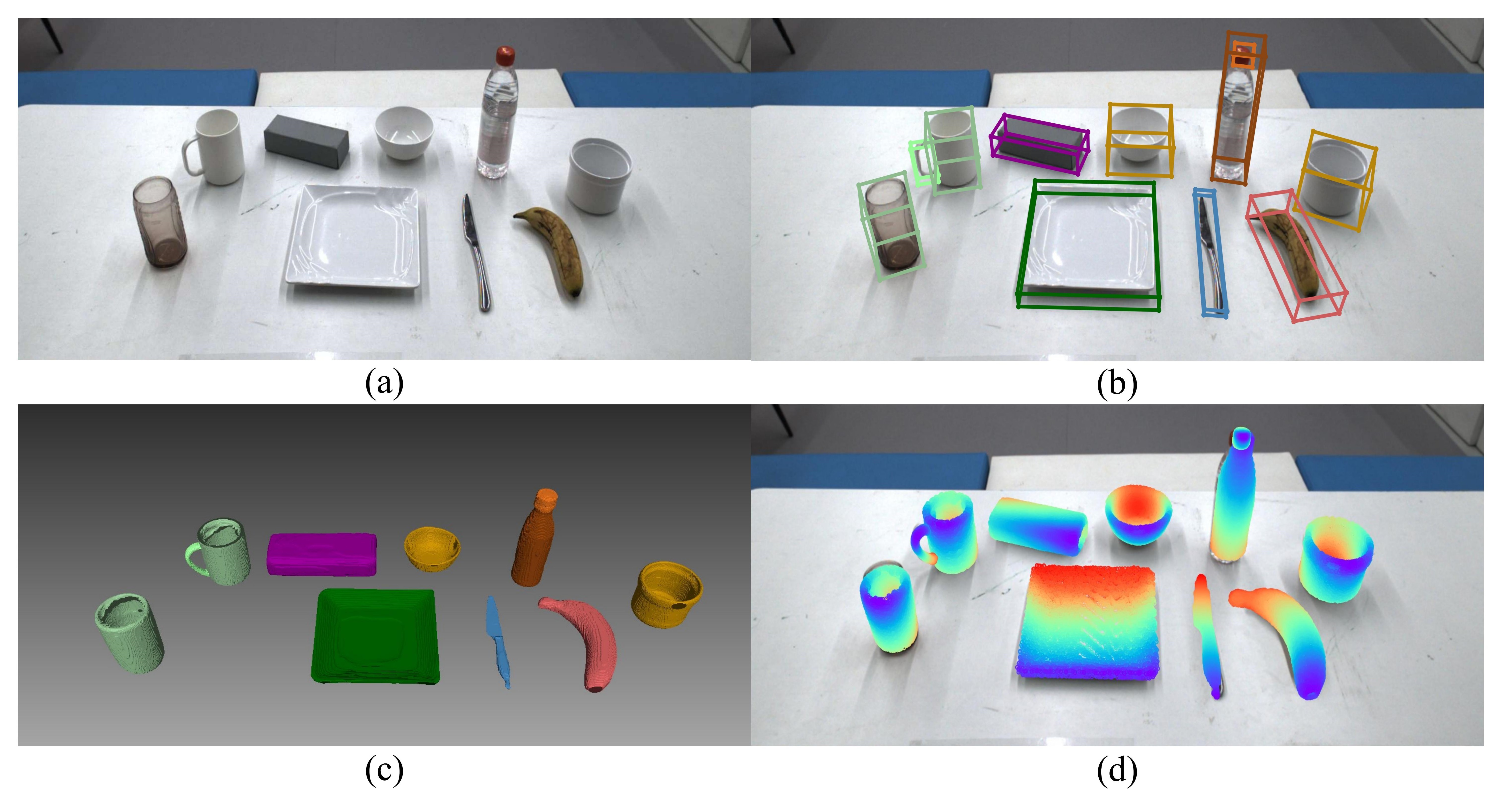}
  \caption{\textbf{Estimations of CODERS on Unseen Objects in Real-world Scenarios.} (a) The left view of the input stereo images with various object surface properties (diffuse, specular, transparent and mixed); (b) Estimated object categories, 6D poses, and sizes; (c) Estimated object shapes; (d) The back-projection of the reconstructed shapes onto the left input image.}
  \label{fig:Result}
\end{figure}

\begin{figure}[tb]
  \centering
  \includegraphics[width=1.0\textwidth]{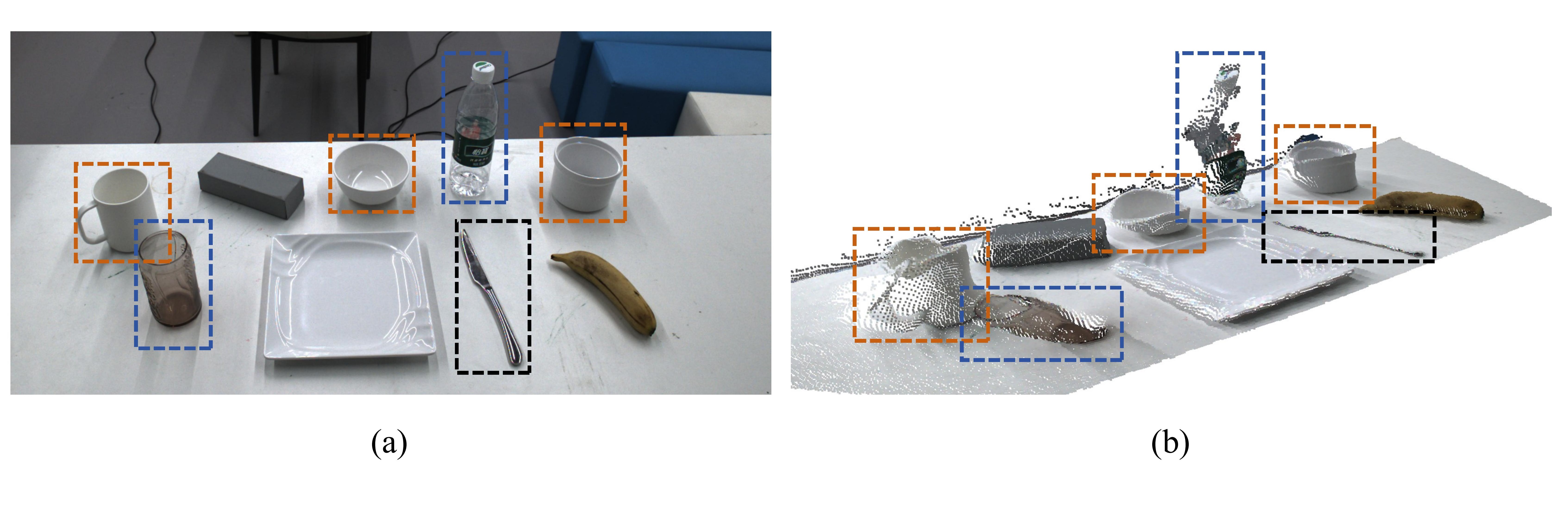}
  \caption{\textbf{Visualization of RGBD measurements in Real-world Experiments} (a) The left view of stereo images. (b) The side view of the obtained colored point cloud.  
  The figure displays the RGB and depth maps of transparent objects, such as the cup and bottle, represented by blue rectangles.
  Polished plastic objects exhibiting high reflection, the bowl and the mug, are indicated by blue rectangles.
  The black rectangles represent steel objects, such as the knife, that exhibit susceptibility to specular reflection.
  Depth measurements of all these objects exhibit both incompleteness and inaccuracies, which limits the performance of RGBD methods. Zoom-in is recommended.
  }
  \label{fig:vis_depth}
\end{figure}

Detecting objects and inferring their 6D poses, shapes and sizes from partial observations are fundamental computer vision tasks for robot manipulation \cite{kollar2022simnet,jiang2021synergies,jiang2022ditto,mees2019self, liu2022akb, Xiang_2020_CVPR} (Fig.~\ref{fig:Result}).
These tasks are known to be challenging due to the diversity of everyday objects in poses, sizes, shapes and surface properties (diffuse, specular, transparent and mixed).
To overcome the inherent scale-ambiguity limitation of monocular methods\cite{wang2021gdr, chen2022epro, sun2022onepose, he2022onepose++, liu2022gen6d}, existing works learn the prior knowledge about the object size by training on large datasets.
However, the obtained accuracy of predicted poses and shapes is unsatisfactory for dexterous manipulation.
RGB-D methods\cite{wang2019normalized,liu2023net, geng2023gapartnet, mo2019partnet,shi2021stablepose}, armed with real-world-scale depth measurements, achieve much better performance on the scale estimations.
The main drawback of RGB-D methods is that depth measurements are missing or imprecise when it comes to objects with specular or transparent surface properties where depths can not be well captured by depth sensors (Fig.~\ref{fig:vis_depth}).
This drawback causes RGB-D methods to face the same scale-ambiguity issue as the monocular approaches.
Inspired by the human binocular vision system, stereo methods seem promising because the real-world scale can be obtained by triangulation with a calibrated stereo baseline.
The key to stereo methods is to answer this question: \textit{how to effectively learn features that are able to handle various surface properties and extract the depth information for the following tasks with the learned features?}

Another focus when deploying algorithms in real-world scenarios is the ability to generalize to unseen objects since object models can usually not be known in advance.
For this reason, discussions on category-level pose estimation and reconstruction have attracted increasing attention in recent years.
Considering objects in specific categories, category-level methods \cite{wang2019normalized, di2022gpv,liu2023net,zhang2022self,liu2020keypose,lin2022category, chen2023stereopose} show their potentials to unseen objects.
Research \cite{chen2023stereopose,liu2020keypose, zhang2022transnet, fang2022transcg} has been conducted on category-level pose estimation for transparent objects using stereo observations. 
However, these methods encounter two primary difficulties when used in robot manipulation. 
The first difficulty is that they commonly employ a two-stage framework, where objects are first detected using detectors, such as Mask-RCNN\cite{he2017mask}, and then their poses as well as shapes are estimated based on the images extracted using 2D bounding boxes from the previous step. 
This pipeline is complex and error-prone as discussed in \cite{heppert2023carto}. 
What's worse, merely using images within detected bounding boxes leads to the potential loss of valuable image information that benefits the following tasks.
The second difficulty is that these methods only address the object pose estimation problem.
Shape reconstructions are not discussed.

To mitigate the challenges above, we present \textbf{CODERS}, a one-stage approach for category-level object detection, pose estimation and reconstruction from stereo images (Fig.~\ref{fig:pipeline}). 
Image depths, indicating the real-world 3D information, are not explicitly computed but implicitly encoded in the learned features (Sect.~\ref{subsec:stereo_matching}).
The obtained 3D-aware features are then used to predict object detections, poses and shapes via a transformer-decoder architecture (Sect.~\ref{subsec:decoder}, Sect.~\ref{subsec:detections_and_pose} and Sect.~\ref{subsec:shapes}).
Unlike existing stereo methods with two stages\cite{chen2023stereopose,liu2020keypose}, our model pipeline is end-to-end without error accumulations between stages or tasks.
To cope with various object surface properties, we train our model with a large simulated dataset covering a diversity of object surface property conditions (Sect.~\ref{subsec:exp_steup}).
To the best of our knowledge, our presented model is the first to use stereo images as input and concurrently estimate object detections, poses and shapes in an end-to-end manner.
Our model significantly outperforms competing stereo methods and demonstrates excellent generalization capability in real-world robot manipulation scenarios where objects are with the same categories and unseen in the training dataset (Sect.~\ref{sec:Experiments}).

In summary, our main contributions are as follows:
\begin{itemize}
\item[$\bullet$] 
We introduce the first end-to-end framework, using stereo images as input, that is able to concurrently estimate object detections, 6D poses and 3D shapes on everyday objects with various surface properties. 
\item[$\bullet$] 
We propose an Implicit Stereo Matching module that implicitly encodes the 3D depth information into the learned image features.
Concatenating this presented module and the following transform-decoder architecture leads to end-to-end multi-task learning required by robot manipulation.
\item[$\bullet$] 
We demonstrate our superior model performance to competing stereo methods in the public TOD dataset \cite{liu2020keypose} and excellent generalization capability in real-world robot applications.
\end{itemize}

\section{Related Work}
\subsection{Category-Level Object Pose Estimation}
Category-level object pose estimation\cite{wang2019normalized, tian2020shape, zhang2024generative, lee2023tta, lunayach2023fsd, chen2023stereopose} focuses on predicting the pose of novel objects within a specific category. 
The object intra-class shape variation is the main challenge of applying the network to novel objects for accurate pose estimation. 
Wang \etal \cite{wang2019normalized} introduced the concept of NOCS (Normalized Object Coordinate Space) representation. 
This representation allows objects to be represented in a normalized canonical space, aiming to mitigate the shape differences encountered during object pose estimation.
Tian \etal \cite{tian2020shape} propose a prior-based framework that explicitly reconstructs the shape of novel objects within the NOCS framework. 
While the NOCS representation and the prior-based framework\cite{ze2022category, weng2021captra, zhang2022self, lin2022category} have gained popularity for category-level object pose estimation, it is vital to acknowledge that most existing methods heavily depend on point cloud data. 
As a result, these methods may not be suitable for transparent object category-level pose estimation. 
Chen \etal \cite{chen2023stereopose}  utilizes stereo images as input and incorporates parallax attention for stereo feature fusion. 
They follow prior-based methods and can estimate the pose of transparent objects on the TOD dataset\cite{liu2020keypose}. 
In contrast to existing methods, our approach does not depend on shape priors and leverages stereo information to achieve category-level pose estimation.

\subsection{3D Shape Reconstruction}
3D object reconstruction\cite{mildenhall2021nerf, kerbl20233d, xie2021toward, liu2023zero, mescheder2019occupancy, irshad2022centersnap} plays a crucial role in the task of 3D object understanding.
Xie \etal \cite{xie2021toward} aims to reconstruct the 3D volume or point cloud of an object using a pair of stereo images. 
They construct a 3D cost volume from stereo features and then decode 3D points using a shape decoder. 
However, this method is limited to instance-level scenes, and as the number of 3D points increases, the training cost becomes huge.
Implicit Neural Representation(INR), which is coordinate-based Multi-Layer Perceptron(MLP)\cite{xie2022neural}, has gained popularity as a method for 3D shape reconstruction. 
Park \etal \cite{park2019deepsdf} represent the Signed Distance Function (SDF) as low-dimensional codes and a corresponding decoder. 
By employing the INR, their method can provide the SDF value for any 3D coordinate of an object, offering a more efficient representation of the object's shape.
Irshad \etal \cite{irshad2022shapo} developed a category-level INR for shape reconstruction, building upon DeepSDF\cite{park2019deepsdf}. 
By leveraging point cloud input, their approach enables the reconstruction of different objects belonging to the same category.
Our approach utilizes SDF to encode objects as implicit shape embeddings and achieves 3D shape reconstruction based on stereo observation in a zero-shot manner.

\section{Method}
\label{sec:Method}

\subsection{Overall Architecture}
\begin{figure}[tb]
  \centering
  \includegraphics[width=1.0\textwidth]{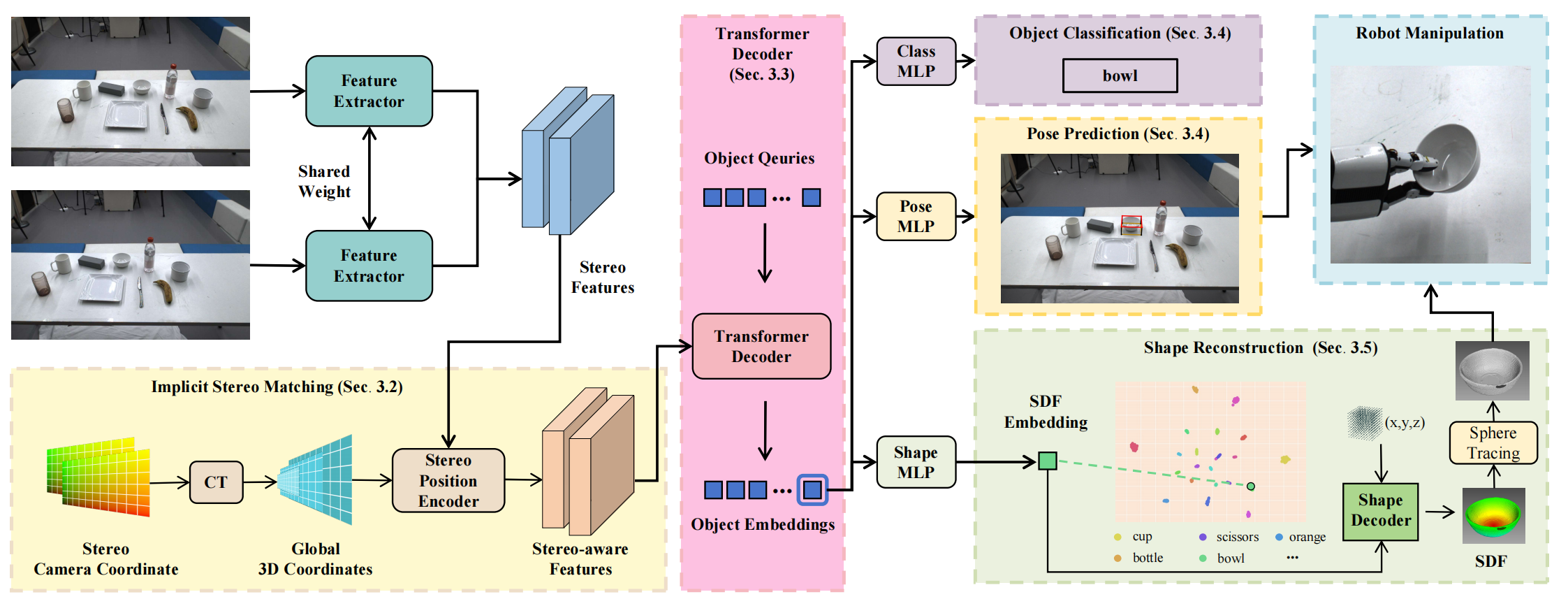}
  \caption{\textbf{Overview of Our Proposed CODERS.} 
  We present a single-stage network capable of processing multiple unknown objects, outputting detections, classes, 6D poses and 3D shapes concurrently. Using stereo images as input, our network generates stereo-aware features for easier alignment in implicit feature space. During the transformer decoder stage, object queries interact with 3D stereo-aware features, yielding object embeddings. These object embeddings are used to infer the category, pose and shape of objects using corresponding modules, which serve as the final output. In the Implicit Stereo Matching module, \textbf{CT} denotes coordinate transformer.}
  \label{fig:pipeline}
\end{figure}

The overview of our method is shown in \Cref{fig:pipeline}.
Our feature extractor network leverages ConvNext\cite{liu2022convnet} and FPN\cite{lin2017feature} to extract 2D stereo features from stereo images. 
In the Implicit Stereo Matching module, we first perform a coordinate transformation from stereo camera coordinates to global 3D space using stereo camera parameters.
Subsequently, we utilize a stereo position encoding network to generate stereo-aware stereo features.
We adopt a transformer decoder to align stereo-aware features with initialized object queries and produce expressive object embeddings.
The resulting object embeddings are utilized to predict object class, pose, and shape with corresponding modules.

\subsection{Implicit Stereo Matching}
\label{subsec:stereo_matching}
In this work, we introduce Implicit Stereo Matching to align stereo features in implicit feature space.
We project stereo camera coordinates to global 3D space to establish the relationship between stereo 2D images.
Inspired by PETR \cite{liu2022petr}, our approach begins by sampling depth values along the axis perpendicular to the image plane. We discretize the camera frustum space to construct 3D meshgrids. 
Then we utilize the reverse 3D projection technique to calculate the corresponding coordinates in global 3D space.

\begin{equation}
  P_{w} = [R,T]K^{-1}P_{c}
  \label{eq:important}
\end{equation}
Where $K \in \mathbb{R}^{4 \times 4}$ denotes the intrinsic parameters of the stereo camera, and $[R,T]$ denotes the transformation matrix from camera space to global 3D space. $P^{w}$ and $P^{c} \in \mathbb{R}^{D \times H \times W \times 4}$ represent the coordinates of points in global 3D space and camera space, respectively.

\begin{figure}[tb]
  \centering
  \includegraphics[width=0.7\textwidth]{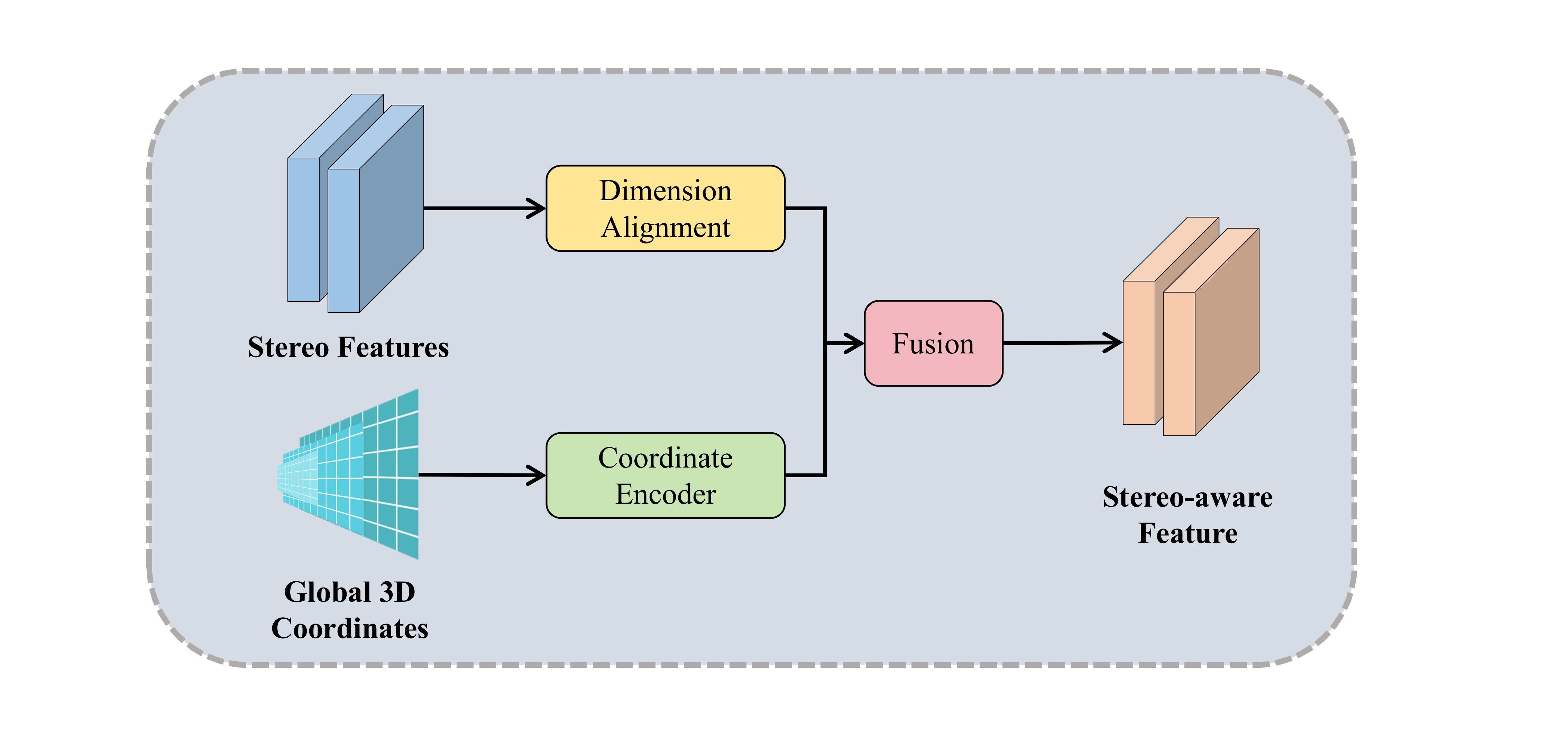}
  \caption{\textbf{Ilustration of Stereo Position Encoding Function.} The stereo features are initially dimension aligned with implicit feature space. Simultaneously, the global 3D coordinates are transformed into stereo 3D position embeddings using coordinate encoder(MLP network). These stereo 3D position embeddings are then fused with the aligned stereo feature to generate stereo-aware features.}
  \label{fig:ism}
\end{figure}

With aligned coordinates, our proposed stereo position encoder obtains the stereo-aware 3D features $F^{3d} = \{ F_i^{3d} \in \mathbb{R}^{C \times H \times W}, i=1,2\}$ by associating the 2D image features $F^{2d} = \{ F_i^{2d} \in \mathbb{R}^{C \times H \times W}, i=1,2\}$ with the 3D position information. Similar to the formulation used in MetaSR\cite{hu2019meta}, we can express the 3D position encoder as follows:

\begin{equation}
  F_{3D} = \psi(F_{2D}, P_w)
  \label{eq:important}
\end{equation}
where $\psi(\cdot)$ represents the stereo position encoding function as illustrated in \Cref{fig:ism}. $F^{2D}, F^{3D} \in \mathbb{R}^{C \times H \times W \times 2}$ are stereo features and stereo-aware features.

\subsection{Transformer Decoder}
\label{subsec:decoder}
We adopt the structure of the standard transformer decoder used in DETR\cite{carion2020end}, which includes L decoder layers.
Each decoder layer consists of a self-attention module for facilitating interaction among object queries, a cross-attention with stereo-aware features to incorporate image features, and a feed forward network (FFN) for updating object queries.
To perform all attention operations, we employ multi-head attention.
Through iterative interactions, the decoder outputs the object embeddings that acquire high-level representations. 
These object embeddings can then be utilized to predict category, pose and shape of the corresponding objects.

\subsection{Object Classification and Pose Prediction}
\label{subsec:detections_and_pose}
In this section, we introduce two branches for object classification and pose prediction. 
We utilize object embeddings along with a corresponding module (in this work, we use an MLP) to perform regression tasks for the object category probability, 6D pose, and size.
To supervise object classification, we employ the focal loss\cite{lin2017focal}. For location and size regression, we use the L1 loss.
To address the issue of discontinuity in rotation prediction, we adopt the approach presented in GDR-Net \cite{wang2021gdr} by predicting a 6-dimensional vector $R_{6d} = [r_1 | r_2]$. 
The rotation matrix $R = [R_{.1} | R_{.2} | R_{.3}]$ can be calculated as follows:

\begin{equation}
\begin{cases}
R_{.1} = \phi(r_1) \\
R_{.3} = \phi(R_{.1} \times r_2) \\
R_{.2} = R_{.3} \times R_{.1} \\
\end{cases}
\end{equation}
where $\phi(\cdot)$ denotes the vector normalization operation.

Similar to DETR\cite{carion2020end}, we use the Hungarian algorithm\cite{kuhn1955hungarian} to perform one-to-one matching between ground truth and predicted values.
The loss for pose prediction can be summarized as follows:

\begin{equation}
L_{pose} = L_{location} + L_{size} + L_{rotation} 
\end{equation}
where $L_{location}$ and $L_{size}$ are $L_1$ losses. $L_{rotation}$ is defined as the average of the $L_1$ loss between predicted rotation $\hat{Rx}$ and ground truth rotation $\bar{Rx}$. 

\subsection{Shape Reconstruction}
\label{subsec:shapes}
Firstly, we implement a category-level shape encoder using Signed Distance Functions (SDF) to generate per-object implicit representations.
We simultaneously train a shape decoder f and the shape embedding z for every object in our dataset. 
With every 3D point $x$, we can easily obtain an approximate Signed Distance Function (SDF) value for the shape.

\begin{equation}
SDF(x) = f(z_i, x)
\end{equation}
where $z_i$ is the corresponding shape embedding for object $i$. 
As our objective is to create a category-level shape encoder, we aim to maximize the dissimilarity between shape embeddings from different categories in the implicit shape space. To achieve this, we incorporate a contrastive loss\cite{khosla2020supervised} during the training of the shape decoder. 
This loss facilitates the quantification of the shared shape characteristics among objects within the same category.

Next, we employ a shape MLP to directly predict the shape embeddings of the objects.
We employ $L_1$ loss for shape embedding regression.

\begin{equation}
L_{shape} = L_1(\hat{z}, \bar{z})
\end{equation}
where $\hat{z}$ represents the predicted shape embedding. 
Combined with object classification and pose prediction losses, our framework can be trained end-to-end.
The total loss is as follows:
\begin{equation}
L_{total} = \lambda_{cls}L_{cls} + \lambda_{pose}L_{pose} + \lambda_{shape}L_{shape}
\end{equation}
The hyperparameters $\lambda_{cls}$, $\lambda_{pose}$, and $\lambda_{shape}$ are utilized to balance the various losses.

\section{Experiments}
\label{sec:Experiments}
\subsection{Experimental Setup}
\label{subsec:exp_steup}
{\bf Datasets} 

To train our network, we generate a large-scale stereo category-level object dataset called \textbf{SS3D}. 
This dataset uses 3D models from OmniObject3D\cite{wu2023omniobject3d}. 
OmniObject3D consists of 6,000 scanned objects with 190 daily categories. 
To build the SS3D dataset, we select a subset of these categories that are suitable for our robot hand to manipulate. 
In \Cref{tab:ss3d}, we provide a list of selected categories along with the corresponding number of objects in each category.

We evaluate our method on the public TOD dataset\cite{liu2020keypose}, which comprises three categories of transparent objects including bottles (3 instances), mugs (7 instances) and cups (2 instances). 
It consists of approximately 36,000 stereo image pairs for 12 different object instances in 10 different environmental backgrounds. 
To conduct category-level experiments, we follow the same settings in \cite{chen2023stereopose} and perform experiments on two different category splits: `mug' and `bottle'. 
We train CODERS on the two categories simultaneously and evaluate our method on novel instances from each category that are not included in the training process.

\begin{table}[tb]
  \caption{\textbf{Overview of Objects in the SS3D Dataset.} 
  In our study, we have chosen a subset of OmniObject3D\cite{wu2023omniobject3d} objects for our training and testing datasets.
  This subset consists of 427 object instances in total, with 363 instances allocated for training and 64 instances(4 instances per category) reserved for testing.} 
  \label{tab:ss3d}
  \centering
  \tabcolsep=0.1cm
  \renewcommand\arraystretch{1.2}
    \begin{tabular}{l|cccccccc}
    \hline
    Category & Banana & Book & Bottle & Bowl     & Carrot & Corn     & Cucumber & Cup   \\ \hline
    Object & 30     & 23   & 33     & 24       & 28     & 31       & 22       & 42    \\ \hline
    Category       & Dish   & Fork & Knife  & LargeBox & Orange & SmallBox & Scissors & Spoon \\  
           \hline
    Object       & 23     & 20   & 22     & 20       & 28     & 37       & 22       & 22    \\ \hline
    \end{tabular}
\end{table}

\noindent
{\bf Implementation Details} 

We utilize ConvNext \cite{liu2022convnet} as the backbone network and FPN \cite{lin2017feature} to aggregate multi-level features.
CODERS is trained with AdamW \cite{loshchilov2017decoupled} using a weight decay of $10^{-2}$.
We initially set the learning rate to $2.0 \times 10^{-4}$ and decay it using a cosine annealing policy\cite{loshchilov2016sgdr}
The loss weights $\lambda_{cls}$, $\lambda_{pose}$, and $\lambda_{shape}$ are assigned as $2$, $6 \times 10^{-2}$, and $2 \times 10^{-2}$, respectively, to achieve a balance among the different losses.
All experiments are trained for 24 epochs on 8 RTX3090 GPUs with a batch size of 8 and tested on a single RTX3090 GPU. 
During inference, no test time augmentation methods are used.

\noindent
{\bf Metrics} 

In line with\cite{wang2019normalized,di2022gpv}, we utilize commonly adopted metrics for evaluating pose prediction. 
These metrics include the mean precision of 3D intersection over union (3DIoU), which enables the joint evaluation of rotation, translation, and size. 
Additionally, we consider the rotation error using thresholds of \{$5^{\circ}$, $10^{\circ}$\} and the translation error using thresholds of \{2 cm, 5 cm, 10 cm\} to evaluate the prediction directly. 
Specifically, a prediction is deemed correct only if it falls within the specified thresholds for both rotation and translation errors.

To assess the quality of reconstruction, we utilize Chamfer distance. 
For this evaluation, we sample 10,000 points from both ground-truth mesh and predicted mesh by our shape reconstruction module. 
Chamfer distance can be calculated as follows:

\begin{equation}
\text{{Chamfer distance}} = \frac{1}{N} \sum_{i=1}^{N} \min_{j} \| \mathbf{x}_i - \mathbf{y}_j \|_2^2 + \frac{1}{M} \sum_{j=1}^{M} \min_{i} \| \mathbf{y}_j - \mathbf{x}_i \|_2^2
\end{equation}
where $\mathbf{x}_i$ represents a point from the ground truth mesh, $\mathbf{y}_j$ represents a point from the predicted mesh, and $N$, $M$ represent the number of sampled points from each mesh, respectively. 
Chamfer distance is an effective metric for quantifying the dissimilarity between two point sets.

\subsection{Pose Estimation Comparison with State-of-the-Art Methods}

In our evaluation, we compare CODERS with state-of-the-art (SOTA) category-level methods on the TOD dataset \cite{liu2020keypose}. 
\Cref{tab:sc} presents the comparative results of our method against other competing approaches. 
Our proposed method demonstrates significant superiority over the competitors across all evaluation metrics. 
SPD \cite{tian2020shape} and SGPA \cite{chen2021sgpa} are RGBD methods. 
We utilize the results reported in StereoPose \cite{chen2023stereopose} for a quick comparison.
KeyPose\cite{liu2020keypose} is a keypoint-based approach that utilizes stereo images to predict key points of category-level objects. StereoPose\cite{chen2023stereopose} is currently the SOTA stereo category-level method on the TOD dataset, achieving better results by predicting back-view NOCS.
Particularly noteworthy is the achievement of CODERS, which attains a \textbf{99.5\%} $3D_{50}$ score in the bottle category, a significant improvement over the 22.2\% reported by the state-of-the-art (SOTA) stereo method. This success is mainly attributed to the effectiveness of our Implicit Stereo Matching module.
Moreover, unlike StereoPose, which relies on object segmentation for pose estimation, CODERS only requires the full image as input, making our method immune to segmentation errors.
We also demonstrate the performance of CODERS on more stringent metrics, which further validates the high capability of our proposed network.

\begin{table}[tb]
  \caption{\textbf{Comparison with State-of-the-Art Methods on TOD Dataset.}
  Here, $3D_{25}$, $3D_{50}$, and $3D_{75}$ refer to the mean precision of 3D intersection over union (3DIoU) with thresholds of 25\%, 50\%, and 75\%, respectively. 
  ${5^{\circ}2 \text{cm}}$ refers to the condition where the error in the object center is limited to less than 2 cm, and the rotation error is restricted to less than 5 degrees.
    CODERS is the SOTA method on the TOD Benchmark by a considerable margin, and our method achieves a \textbf{99.5\%} $3D_{50}$ score in the bottle category, marking a significant improvement from the 22.2\% of previous method.
  The larger the value, the better the performance.}
  \label{tab:sc}
  \centering
  \tabcolsep=0.05cm
  \renewcommand\arraystretch{1.2}
  \begin{tabular}{l|c|c|c|c|c|c|c|c|c|c|c|c}
    \hline
    \multirow{2}{*}{Method} & \multicolumn{6}{c|}{Bottle} & \multicolumn{6}{c}{Mug}\\   \cline{2-13}
    \multirow{2}{*}{} & $3D_{25}$ & $3D_{50}$ & $3D_{75}$ & $\makecell[c]{5^{\circ}\\2cm}$ & $\makecell[c]{10^{\circ}\\5cm}$ & $\makecell[c]{10^{\circ}\\10cm}$ & $3D_{25}$ & $3D_{50}$ & $3D_{75}$ & $\makecell[c]{5^{\circ}\\2cm}$ & $\makecell[c]{10^{\circ}\\5cm}$ & $\makecell[c]{10^{\circ}\\10cm}$ \\
    \hline
    SPD\cite{tian2020shape} & 44.5 & 7.4 & - & - & 11.5 & 17.8 & 63.6 & 19.7 & - & - & 2.3 & 4.2 \\
    SGPA\cite{chen2021sgpa} & 46.9 & 9.6 & - & - & 13.3 & 22.5 & 64.6 & 19.6 & - & - & 2.8 & 5.1 \\
    KeyPose\cite{liu2020keypose} & - & - & - & - & 52.7 & 62.3 & - & - & - & - & 24.6 & 25.1 \\
    StereoPose\cite{chen2023stereopose} & 85.4 & 22.2 & - & - & 57.8 & 70.3 & 97.9 & 77.4 & - & - & 34.4 & 38.2 \\
    \hline
    Ours & \textbf{100} & \textbf{99.5} & \textbf{31.5} & \textbf{73.4} & \textbf{99.8} & \textbf{99.8} & \textbf{100} & \textbf{100} & \textbf{92.8} & \textbf{56.2} & \textbf{81.9} & \textbf{81.9} \\
    \hline
  \end{tabular}
\end{table}

\subsection{Reconstruction Comparison with State-of-the-Art Methods}

In our study, we compare CODERS with StereoPoints\cite{xie2021toward}, Zero123\cite{liu2023zero} and TripoSR\cite{tochilkin2024triposr} on five unseen objects with scanned ground truth. 
To the best of our knowledge, StereoPoints is the only method that performs stereo shape reconstruction from a single view. 
However, it is unable to reconstruct unseen objects directly. 
Therefore, we finetune the results of StereoPoints on the five objects. 
On the other hand, Zero123 and TripoSR are zero-shot 3D shape reconstruction method that can generate various objects using single-view images. 
To ensure a fair comparison, we adopt the same setting as StereoPoints, Zero123 and TripoSR, which requires object-centric images without background, while CODERS utilizes the entire image. 
We adopt Chamfer distance to measure the similarity between the ground truth mesh and the extracted mesh, and the output distances are multiplied by 100. 
The results of our comparison are presented in \Cref{tab:rc}.

CODERS demonstrates superior performance compared to Zero123 and TripoSR for all five objects, indicating that our method can generate higher-quality meshes for category-level unseen objects. 
This improvement is attributed to our category-level shape decoder. 
StereoPoints outperforms our method in the cup and knife categories because it specializes in instance-level reconstruction. 
Their network is trained and tested on the same objects, and cannot generalize to unseen objects.
CODERS achieves comparable results with StereoPoints and even outperforms it on bottle, bowl, and soup objects. 
This highlights the efficiency of our proposed category-contrastive shape embedding encoder, which enhances the variability within shape embeddings.
Furthermore, we conducted a comparison with our method that does not include the category-contrastive shape embedding encoder. 
The results further validate the effectiveness of our proposed encoder in improving the performance of CODERS.

\begin{table}[tb]
  \caption{\textbf{Reconstruction Comparisons.}
    Chamfer distance (CD) is a metric used to evaluate the similarity of two point clouds. 
    `Ours-nocontr' denotes our results obtained without utilizing contrastive loss in the shape encoder.
    Our method achieves results on par with Stereo2Point, despite the latter being trained and tested on the same object.
    A lower value of CD indicates better performance.
  }
  \label{tab:rc}
    \centering
      \tabcolsep=0.2cm
    \renewcommand\arraystretch{1.2}
    \begin{tabular}{l|c|c|c|c|c}
    \hline
    \multirow{2}{*}{Method}  & bottle & bowl & cup  & knife & soup \\
    \cline{2-6}
    \multirow{2}{*}{}    & \multicolumn{5}{c}{CD($\downarrow$)}              \\
            \hline
    Zero123 \cite{liu2023zero}          & 1.266             & 1.042             & 0.685             & 0.614            & 1.524 \\
    TripoSR \cite{tochilkin2024triposr} & 1.478             & 0.842             & 0.908            & 1.008            & 2.426  \\
    Stereo2Point \cite{xie2021toward}   & 0.700             & 0.509             & \textbf{0.563}    & \textbf{0.140}   & 0.695 \\
    \hline
    Ours-nocontr                        & 1.411             & 0.516             & 0.612             & 0.272            & 0.482 \\
    Ours                                & \textbf{0.542}    & \textbf{0.352}    & 0.593             & 0.211            & \textbf{0.458} \\
    \hline
\end{tabular}
\end{table}

\subsection{Evaluation on Our Generated SS3D Dataset}
We evaluate our method on the SS3D test dataset which contains 64 unseen objects with various materials including specular and transparent ones. 
The results in \Cref{tab:ESS3D} show the generalization ability of CODERS to various sizes, shapes and materials.
Our proposed network can handle most categories within the SS3D dataset, including books, bottles, corn, and dishes, benefiting from the stable depth information provided by our Implicit Stereo Matching module. 
However, our method struggles with larger boxes, primarily due to the giant intra-class size variation and size distribution outliers from other categories.
To address this issue, we require additional training data and a more balanced category distribution.

\begin{table}[tb]
    \centering
    \scriptsize
    \label{tab:ESS3D}
    \caption{\textbf{Results on SS3D Test Dataset.} 
    We evaluate CODERS on the SS3D test dataset with 16 categories of unseen objects. 
    In the notation, \textbf{CA.} stands for Category, while \textbf{BA.} represents Banana. 
    The first row corresponds to categories in reference to \Cref{tab:ss3d}.
    Our method can manage objects across all 16 categories, encompassing various sizes, shapes and materials, demonstrating the generalization capability of our stereo framework.
    }
    \tabcolsep=0.05cm
    \renewcommand\arraystretch{1.5}
    \begin{tabular}{l|cccccccccccccccc}
    \hline
    \textbf{CA.} & \textbf{BA.}  & \textbf{BO.}  & \textbf{BOT.} & \textbf{BOW.} & \textbf{CA.}  & \textbf{CO.}  & \textbf{CU.}  & \textbf{CUP}  & \textbf{DI.}  & \textbf{FO.}  & \textbf{KN.}  & \textbf{LA.}  & \textbf{OR.}  & \textbf{SM.}  & \textbf{SC.}  & \textbf{SP.}  \\
    \hline
    $3D_{50}$          & 72 & 90 & 71 & 59 & 42 & 89 & 34 & 62 & 88 & 54 & 71 & 52 & 35 & 66 & 48 & 42 \\
    $3D_{75}$          & 26 & 51 & 16 & 20 & 11 & 45 & 9  & 16 & 37 & 15 & 12 & 9  & 6  & 32 & 18 & 10 \\
    $5^{\circ} 5cm$    & 47 & 65 & 80 & 58 & 51 & 74 & 38 & 72 & 64 & 58 & 61 & 16 & 56 & 32 & 51 & 33 \\
    $5^{\circ} 2cm$    & 26 & 43 & 33 & 23 & 22 & 48 & 20 & 33 & 20 & 33 & 33 & 5  & 25 & 24 & 29 & 20 \\
    \hline
    \end{tabular}
\end{table}

\subsection{Ablation Study}

We present several ablation experiments on the TOD dataset\cite{liu2020keypose}.

\noindent
{\bf Effectiveness of Implicit Stereo Matching}

\begin{table}[tb]
  \caption{\textbf{Ablation Study Results.} 
  All ablation studies are conducted on the TOD dataset using the same training strategy.
  The first experiment highlights the role of our proposed Implicit Stereo Matching module in the fusion of stereo features.
  Moreover, the stereo framework is capable of achieving significantly more accurate results than its monocular counterpart.
  We adopt six decoder layers as a balance between accuracy and computational cost.
  }
  \label{tab:AS}
    \centering
      \tabcolsep=0.2cm
    \renewcommand\arraystretch{1.2}
    \begin{tabular}{l|c|c|c|c}
    \hline
    \multirow{2}{*}{Method} & \multicolumn{2}{c|}{Bottle} & \multicolumn{2}{c}{Mug} \\
    \cline{2-5}
    \multirow{2}{*}{} & $3D_{75}$ & $5^{\circ} 2cm$ & $3D_{75}$ & $5^{\circ} 2cm$ \\
    \hline
    No position embedding                   & 25.8      & 53.5      & 88.4         & 52.5  \\
    2D position embedding                   & 28.6      & 70.8      & 91.7         & 55.3  \\
    Stereo-aware position embedding         & \textbf{31.5}      & \textbf{73.4}      & \textbf{92.8}         & \textbf{56.2}  \\
    \hline
    Monocular                               & 16.1      & 48.0      & 86.1         & 54.7  \\
    Stereo                                  & \textbf{31.5}      & \textbf{73.4}      & \textbf{92.8}         & \textbf{56.2}  \\
    \hline
    1 deocoder layer                        & 24.3      & 29.6      & 66.6         & 26.3  \\
    3 deocoder layer                        & 30.9      & 66.9      & 91.3         & 54.5  \\
    6 deocoder layer                        & 31.5      & 73.4      & 92.8         & 56.2  \\
    8 deocoder layer                        & \textbf{31.7}      & \textbf{73.5}      & \textbf{93.2}         & \textbf{56.8}  \\
    \hline
\end{tabular}
\end{table}

\begin{figure}[tb]
  \centering
  \includegraphics[width=1.0\textwidth]{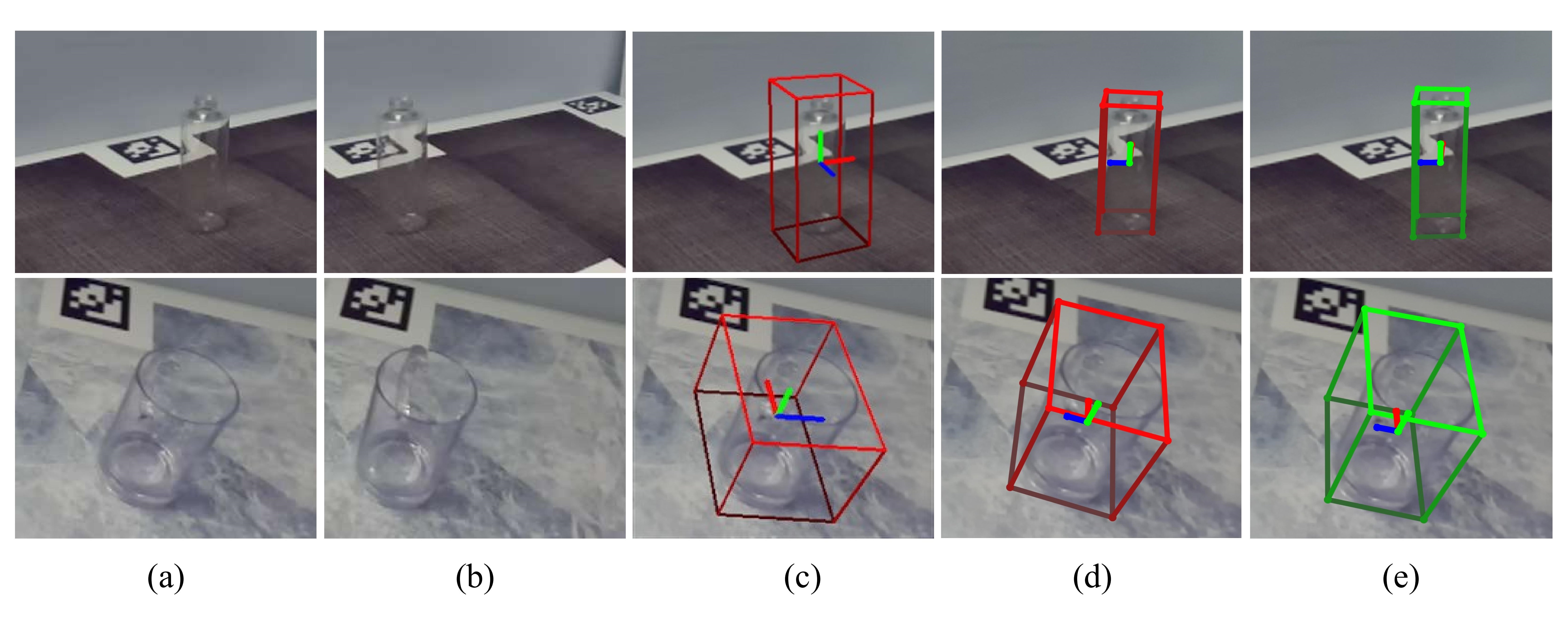}
  \caption{\textbf{Visualization of StereoPose and CODERS on TOD Dataset.} 
  (a) Left view image. (b) Right view image. (c) Results of StereoPose. (d) Results of CODERS. (e) Ground-truths.
  Our method surpasses stereopose in predicting location, size, and rotation. 
  }
  \label{fig:vis_with_stereopose}
\end{figure}

We have conducted experiments to assess the effectiveness of our Implicit Stereo Matching module by comparing it with two alternative approaches: 2D position embedding and no position embedding. 
Specifically, we modified the coordinate transformer and eliminated depth information to construct the 2D position embedding. 
For the no position embedding approach, we adopt a zero embedding.
As demonstrated in \Cref{tab:AS}, the results clearly indicate that our stereo-aware shape embedding approach outperforms alternative methods. 
This finding reinforces the significance of our Implicit Stereo Matching module, highlighting its importance in achieving notable accuracy.

\noindent
{\bf Stereo or Monocular}

We have done experiments to investigate the advantages of using the stereo camera over the monocular camera. 
To carry out this comparison, we separately remove the left and right images to train two models with identical 3D position embeddings. Then, we fuse the results of the two models to achieve better pose estimation results.
The results, as shown in \Cref{tab:AS}, clearly indicate that the stereo approach outperforms the monocular approach by a significant margin, particularly in the case of bottles.
One key reason for this performance difference is the limitation of monocular cameras in providing reliable depth information, especially when objects vary in size such as bottles.
On the other hand, using stereo images with their stable depth information enables our model to accurately estimate the depth and size of objects, even when dealing with objects of unknown size.

\begin{figure}[tb]
  \centering
  \includegraphics[width=1.0\textwidth]{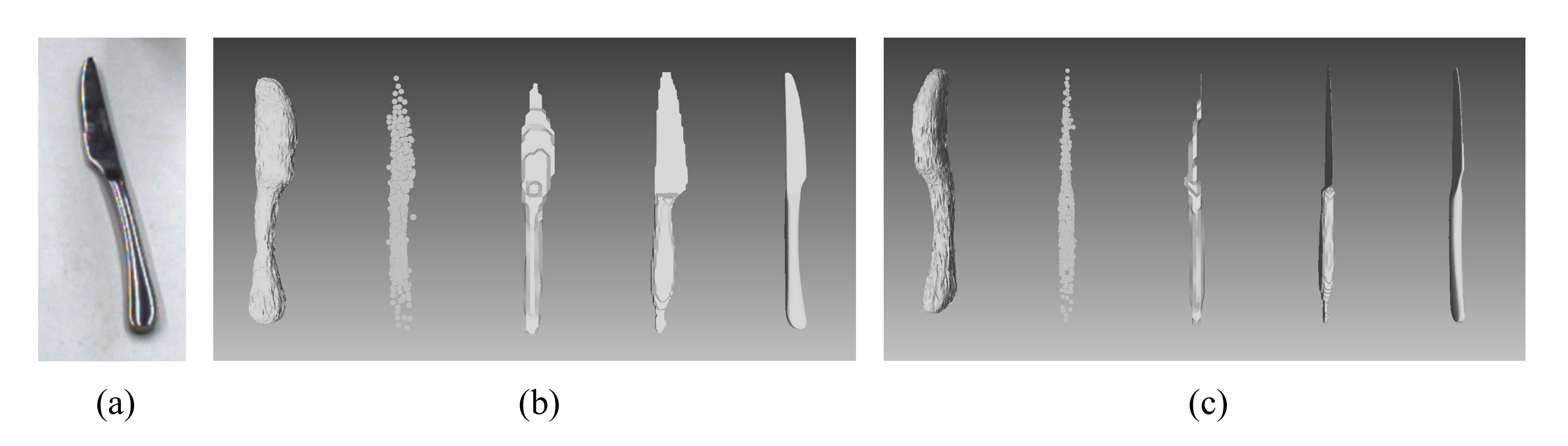}
  \caption{\textbf{Visualization of Reconstruction Results on Knife.} (a) Input image. (b) Font View. (c) Side View. 
  For each view, from left to right: Zero123, StereoPoints(point cloud), CODERS(no contrastive loss), CODERS and Ground-truth.
  Our approach generates meshes with quality comparable quality as instance-level methods (StereoPoints) and can reconstruct blade shapes.
  }
  \label{fig:vis_rc}
\end{figure}

\noindent
{\bf Number of Decoder layers}

In \Cref{tab:AS}, we present the results obtained using different numbers of decoder layers. 
It is evident that when using only one decoder layer, the accuracy is considerably low. 
This shows the importance of the transformer decoder in generating high-quality object embeddings. 
However, it is worth noting that the overall accuracy can be further improved by increasing the number of decoder layers. 
As a trade-off between accuracy and memory usage, we adopt six decoder layers for our network.

\noindent
{\bf Multiple Heads}
For multiple heads ablations, poses/3D bounding boxes are evaluated w. and w.t. the reconstruction heads. 
Reconstructions are evaluated w. and w.t. the pose heads (This is done by weighting the pose head with a normal and very small value).
Results are shown in \Cref{tab:MH}.
The results suggest that multiple heads can enhance both pose and shape performance.

\begin{table}[tb]
    \centering
    \vspace{4pt}
    \tabcolsep=0.1cm
    \renewcommand\arraystretch{1}
    \begin{tabular}{c|c|c|c|c|c|c|c}
    \hline
    \multirow{2}{*}{Pose} & \multirow{2}{*}{Shape} & \multicolumn{2}{c|}{Bottle} & \multicolumn{2}{c|}{Mug} & Bottle & Mug\\
    \cline{3-8}
    \multirow{2}{*}{} & \multirow{2}{*}{} & $3D_{75}$ & $5^{\circ} 2cm$ & $3D_{75}$ & $5^{\circ} 2cm$ & \multicolumn{2}{c}{CD($\downarrow$)}\\
    \hline
    \ding{51}  & \ding{55}    & 24.1      & 61.1      & 86.2         & 47.8  & - & - \\
    \ding{55}  & \ding{51}    & -      & -      & -         & -     & 0.223   & 0.193 \\
    \ding{51}  & \ding{51}    & \textbf{31.5}  & \textbf{73.4} & \textbf{92.8}  & \textbf{56.2} & \textbf{0.201} & \textbf{0.153} \\
    \hline
\end{tabular}
  \caption{\textbf{Ablations on multiple heads.} Both pose and shape heads contribute to the performance increasement for the pose estimation and reconstruction tasks.}
    \label{tab:MH}
\end{table}

\subsection{Qualitative Results}
\Cref{fig:vis_with_stereopose} presents qualitative results of pose estimation on the TOD dataset. 
Our method outperforms StereoPose in terms of object depth, size, and pose estimation accuracy. 
We also present the visualization of the reconstruction results on the knife using the methods mentioned in \Cref{tab:rc}. 
As depicted in \Cref{fig:vis_rc}, CODERS is capable of generating high-quality meshes with only single-view stereo images as input.

\section{Conclusion}
In this work, we introduced CODERS, a single-stage approach for category-level object detection, pose estimation, and reconstruction from stereo images. 
Equipped with stereo images, our method could adeptly handle everyday objects made from diverse materials.
To effectively fuse stereo features, we employed an Implicit Stereo Matching module that aligned them within the implicit feature space.
To construct a single-stage, multi-task pipeline, we encoded the objects into object embeddings, enabling CODERS to simultaneously output category, pose, and shape information with a single forward pass.
Our method achieved state-of-the-art (SOTA) performance on the TOD dataset with a substantial margin. 
The principal limitation of CODERS lies in its inference speed. 
Running on an NVIDIA RTX3090 GPU, our method achieves a rate of 3 Hz, which falls short of the criteria for real-time performance.
Additionally, we plan to release our dataset SS3D as well as code in the future. 
We hope that our approach can enhance the attention toward stereo vision in multi-task settings and serve as a baseline method for subsequent studies.

\section*{Acknowledgement}
\begin{sloppypar}
This research was in part supported by Tencent robotics X and National Key Research and Development Program of China (Project No. 2022YFB36066), in part by the Shenzhen Science and Technology Project under Grant (JCYJ20220818101001004).
\end{sloppypar}

\appendix

\section{Experiment Details}
\subsection{Network Details}
In this section, we will provide a detailed introduction to the specific aspects of our network. As shown in \Cref{tab:sc}, our method utilizes ConvNext-B\cite{liu2022convnet} as image backbone. We output the last two layers of the backbone and adopt FPN\cite{lin2017feature} to further aggregate multi-dimensional information. Then our Implicit Stereo Matching produces 3D stereo embeddings with the same dimension as stereo features. We simply sum the 3D stereo features with 3D stereo embeddings as discussed in our paper.
In the transformer decoder stage, we use 150 object queries to generate object embeddings. We use a 64-dimensional vector to represent the shape of objects.

\begin{table}[H]
\caption{\textbf{Network Details of CODERS.}}
\centering
\label{tab:sc}
\renewcommand{\arraystretch}{1.5}
\setlength{\tabcolsep}{10pt}
\begin{tabular}{|c|c|}
\hline
Layers        & Dimensions                                                                      \\
\hline
Stereo Images & 2 $\times$ 3 $\times$ 608 $\times$ 960                                                              \\
\hline
\begin{tabular}[c]{@{}c@{}}Backbone\\ (ConvNext-B)\end{tabular}      & \begin{tabular}[c]{@{}l@{}}2 $\times$ 512 $\times$ 38 $\times$ 60\\ 2 $\times$ 1024 $\times$ 19 $\times$ 30\end{tabular} \\
\hline
\begin{tabular}[c]{@{}c@{}}Neck\\ (FPN)\end{tabular}           & 2 $\times$ 256 $\times$ 38 $\times$ 60                                                             \\
\hline
3D Stereo Embeddings & 2 $\times$ 256 $\times$ 38 $\times$ 60                                                              \\
\hline
Stereo-aware Features & 2 $\times$ 256 $\times$ 38 $\times$ 60                                                              \\
\hline
Object Queries & 150 $\times$ 256                                                            \\
\hline
Object Embedding & 256                                                            \\
\hline
Shape Embedding & 64                                                            \\
\hline
\end{tabular}
\end{table}

Our transformer decoder contains six decoder layers. In each layer, we process object queries with the order of self-attention, norm, FFN, cross-attention, norm, FFN. The details of the decoder layer are shown in \Cref{tab:dl}. 
Inspired by DETR\cite{carion2020end}, We use multi-head attention for all attention operations.

\begin{table}[tb]
\caption{\textbf{Network Details of Decoder Layer.}}
\centering
\label{tab:dl}
\renewcommand{\arraystretch}{1.5}
\setlength{\tabcolsep}{10pt}
\begin{tabular}{|c|c|c|}
\hline
Layer           & Q                                                           & KV                                                                   \\
\hline
Self-attention  & \begin{tabular}[c]{@{}c@{}}Queries\\ 150 $\times$ 256\end{tabular} & \begin{tabular}[c]{@{}c@{}}Queries\\ 150 $\times$ 256\end{tabular}          \\
\hline
Cross-attention & \begin{tabular}[c]{@{}c@{}}Queries\\ 150 $\times$ 256\end{tabular} & \begin{tabular}[c]{@{}c@{}}Features\\ 2 $\times$ 38 $\times$ 60 $\times$ 256 \end{tabular} \\
\hline
\end{tabular}
\end{table}

\subsection{Experiment Settings}
The TOD\cite{liu2020keypose} dataset provides stereo images with a resolution of 720 $\times$ 1280 pixels. To ensure consistency with our SS3D dataset, we randomly resize and crop the input images to 600 $\times$ 960 pixels.

To maintain uniformity, we train cups and mugs in the same category, labeled as "cup" during the training process.

The Origin TOD dataset only offers key point annotations. To determine the object scale, we measure it using CAD models. Based on these measurements, we generate 6D pose annotations utilizing the provided key points.

Our real-world data is captured at a resolution of 1200 $\times$ 1920 pixels. However, for our purpose, we resize the input images to 600 $\times$ 960 pixels.

For the purpose of reconstruction comparison, we selected zero123 and TripoSR. To meet their requirements, we preprocess our input images to be object-centric and free from background interference.

In the ablation study, we utilize the TOD dataset and follow the same settings as discussed above.

\begin{figure}[tb]
  \centering
  \includegraphics[width=1.0\textwidth]{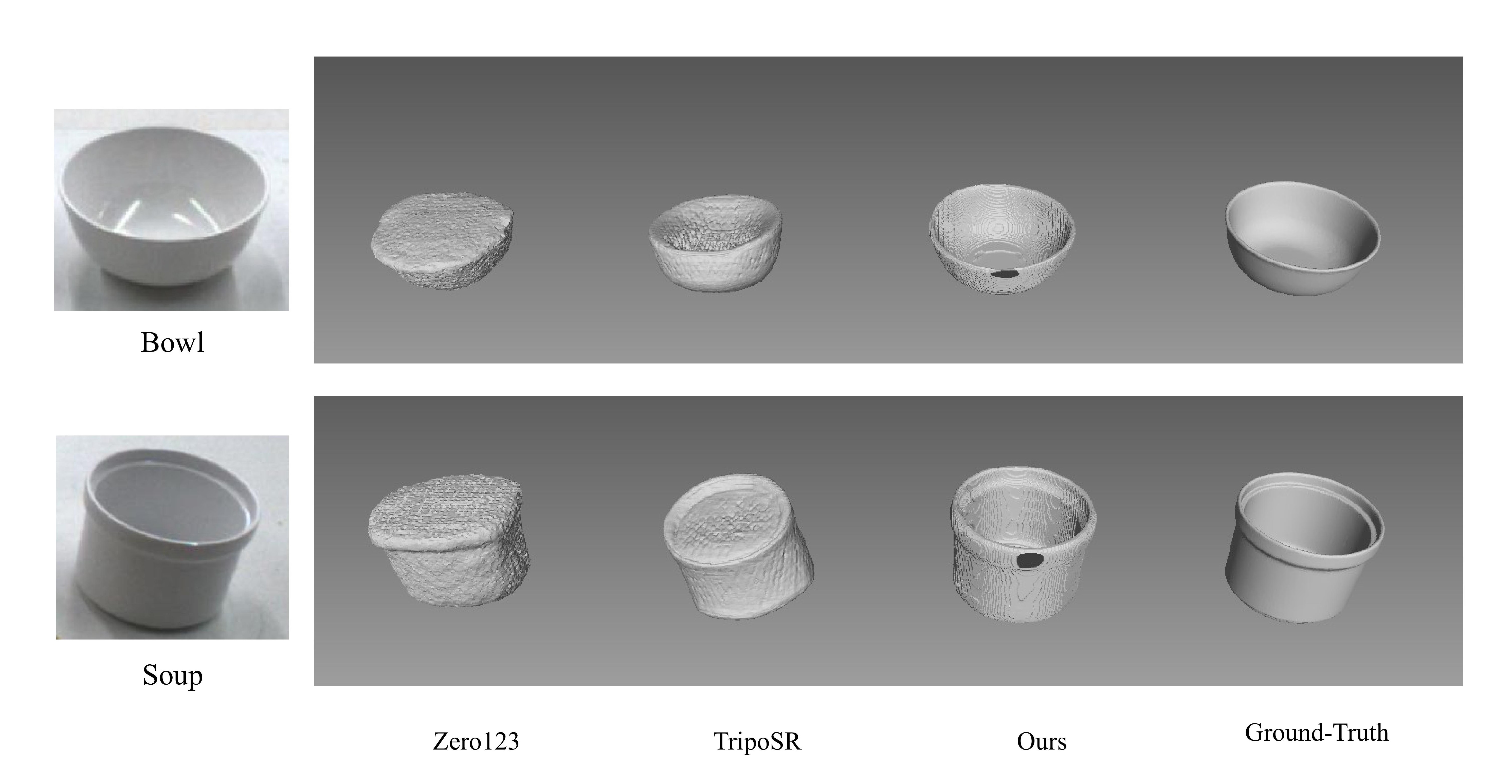}
  \caption{\textbf{Qualitative Results of Reconstruction}
    Our approach generates meshes with high quality and can reconstruct blade shapes.
  }
  \label{fig:vis_rc}
\end{figure}

\section{Qualitative Results of Reconstruction}

We do more comparison for CODERS with Zero123\cite{liu2023zero} and TripoSR\cite{tochilkin2024triposr} on real-world data. 
To ensure a fair comparison, we adopt the same setting as Zero123 and TripoSR, which require object-centric images without background, while CODERS utilizes the entire image. 
As depicted in \Cref{fig:vis_rc}, CODERS is capable of generating high-quality meshes with only single-view stereo images as input.

\begin{figure}[H]
  \centering
  \includegraphics[width=0.8\textwidth]{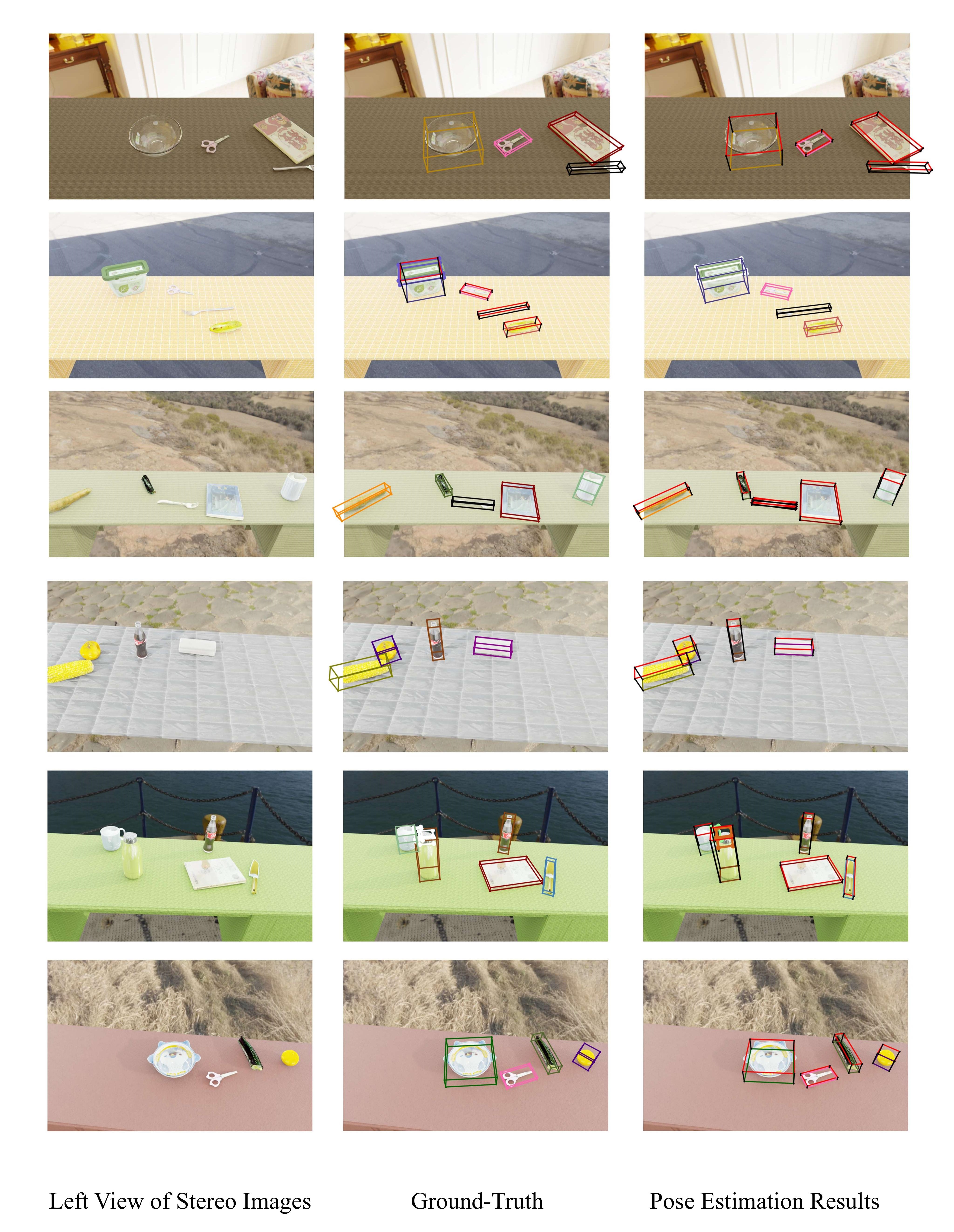}
  \caption{\textbf{Qualitative Results on SS3D Test Dataset}
    The bottom color of the 3D bounding boxes represents the object category. Our method can handle all 16 object categories using a single model. Importantly, these objects have varying surface properties including specular, transparent, and diffuse.
  }
  \label{fig:vis_ss3d}
\end{figure}

\section{Qualitative Results on SS3D Test Dataset}

We generate a large-scale stereo category-level object dataset called \textbf{SS3D}. 
To build the SS3D dataset, we select 427 objects from OmniObject3D\cite{wu2023omniobject3d} and reserve 64 objects from 16 categories for testing.
The results are shown in \Cref{fig:vis_ss3d}.
Our method demonstrates excellent generalization ability for scenes and objects.

\section{Failure Cases}

\begin{figure}[tb]
  \centering
  \includegraphics[width=0.8\textwidth]{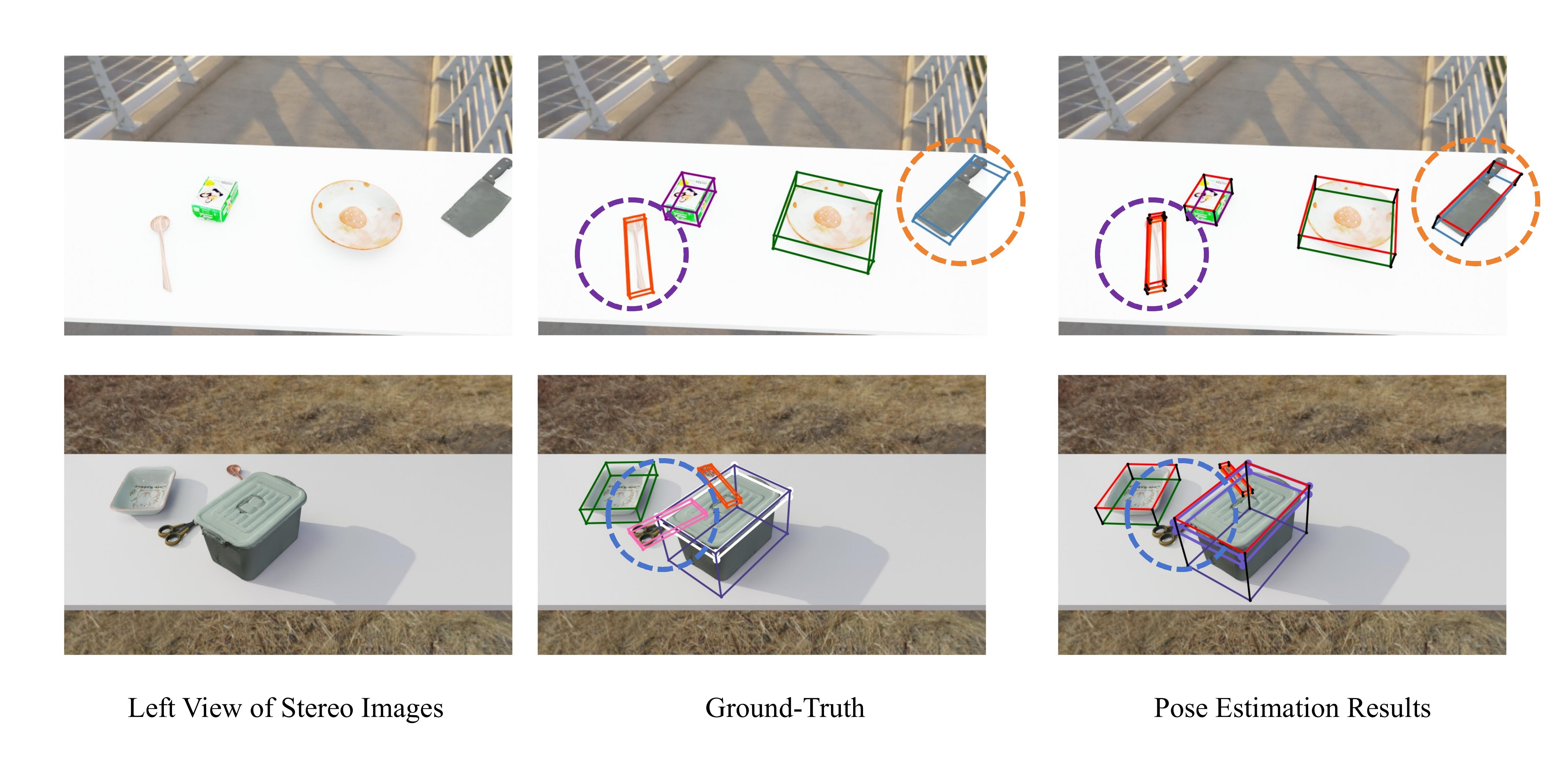}
  \caption{\textbf{Failure Cases}
    The purple circle indicates a wooden spoon with low confidence. The orange circle represents a knife with an outlier scale. The blue circle denotes a pair of scissors that is occluded. These issues are common problems in the field of computer vision, which require further exploration.
  }
  \label{fig:vis_fail}
\end{figure}

In this section, we show some failure cases of CODERS on unseen objects see \Cref{fig:vis_fail}.
Objects with outlier scales, strange materials, and occlusions remain significant issues.
These issues are common problems in the field of computer vision, which require further exploration.

\par\vfill\par


%
%

\bibliographystyle{splncs04}
\bibliography{main}

\begin{thebibliography}{10}
\providecommand{\url}[1]{\texttt{#1}}
\providecommand{\urlprefix}{URL }
\providecommand{\doi}[1]{https://doi.org/#1}

\bibitem{carion2020end}
Carion, N., Massa, F., Synnaeve, G., Usunier, N., Kirillov, A., Zagoruyko, S.: End-to-end object detection with transformers. In: European conference on computer vision. pp. 213--229. Springer (2020)

\bibitem{chen2022epro}
Chen, H., Wang, P., Wang, F., Tian, W., Xiong, L., Li, H.: Epro-pnp: Generalized end-to-end probabilistic perspective-n-points for monocular object pose estimation. In: Proceedings of the IEEE/CVF Conference on Computer Vision and Pattern Recognition. pp. 2781--2790 (2022)

\bibitem{chen2021sgpa}
Chen, K., Dou, Q.: Sgpa: Structure-guided prior adaptation for category-level 6d object pose estimation. In: Proceedings of the IEEE/CVF International Conference on Computer Vision. pp. 2773--2782 (2021)

\bibitem{chen2023stereopose}
Chen, K., James, S., Sui, C., Liu, Y.H., Abbeel, P., Dou, Q.: Stereopose: Category-level 6d transparent object pose estimation from stereo images via back-view nocs. In: 2023 IEEE International Conference on Robotics and Automation (ICRA). pp. 2855--2861. IEEE (2023)

\bibitem{di2022gpv}
Di, Y., Zhang, R., Lou, Z., Manhardt, F., Ji, X., Navab, N., Tombari, F.: Gpv-pose: Category-level object pose estimation via geometry-guided point-wise voting. In: Proceedings of the IEEE/CVF Conference on Computer Vision and Pattern Recognition. pp. 6781--6791 (2022)

\bibitem{fang2022transcg}
Fang, H., Fang, H.S., Xu, S., Lu, C.: Transcg: A large-scale real-world dataset for transparent object depth completion and a grasping baseline. IEEE Robotics and Automation Letters  \textbf{7}(3),  7383--7390 (2022)

\bibitem{geng2023gapartnet}
Geng, H., Xu, H., Zhao, C., Xu, C., Yi, L., Huang, S., Wang, H.: Gapartnet: Cross-category domain-generalizable object perception and manipulation via generalizable and actionable parts. In: Proceedings of the IEEE/CVF Conference on Computer Vision and Pattern Recognition. pp. 7081--7091 (2023)

\bibitem{he2017mask}
He, K., Gkioxari, G., Doll{\'a}r, P., Girshick, R.: Mask r-cnn. In: Proceedings of the IEEE international conference on computer vision. pp. 2961--2969 (2017)

\bibitem{he2022onepose++}
He, X., Sun, J., Wang, Y., Huang, D., Bao, H., Zhou, X.: Onepose++: Keypoint-free one-shot object pose estimation without cad models. Advances in Neural Information Processing Systems  \textbf{35},  35103--35115 (2022)

\bibitem{heppert2023carto}
Heppert, N., Irshad, M.Z., Zakharov, S., Liu, K., Ambrus, R.A., Bohg, J., Valada, A., Kollar, T.: Carto: Category and joint agnostic reconstruction of articulated objects. In: Proceedings of the IEEE/CVF Conference on Computer Vision and Pattern Recognition. pp. 21201--21210 (2023)

\bibitem{hu2019meta}
Hu, X., Mu, H., Zhang, X., Wang, Z., Tan, T., Sun, J.: Meta-sr: A magnification-arbitrary network for super-resolution. In: Proceedings of the IEEE/CVF conference on computer vision and pattern recognition. pp. 1575--1584 (2019)

\bibitem{irshad2022centersnap}
Irshad, M.Z., Kollar, T., Laskey, M., Stone, K., Kira, Z.: Centersnap: Single-shot multi-object 3d shape reconstruction and categorical 6d pose and size estimation. In: 2022 International Conference on Robotics and Automation (ICRA). pp. 10632--10640. IEEE (2022)

\bibitem{irshad2022shapo}
Irshad, M.Z., Zakharov, S., Ambrus, R., Kollar, T., Kira, Z., Gaidon, A.: Shapo: Implicit representations for multi-object shape, appearance, and pose optimization. In: European Conference on Computer Vision. pp. 275--292. Springer (2022)

\bibitem{jiang2022ditto}
Jiang, Z., Hsu, C.C., Zhu, Y.: Ditto: Building digital twins of articulated objects from interaction. In: Proceedings of the IEEE/CVF Conference on Computer Vision and Pattern Recognition. pp. 5616--5626 (2022)

\bibitem{jiang2021synergies}
Jiang, Z., Zhu, Y., Svetlik, M., Fang, K., Zhu, Y.: Synergies between affordance and geometry: 6-dof grasp detection via implicit representations. arXiv preprint arXiv:2104.01542  (2021)

\bibitem{kerbl20233d}
Kerbl, B., Kopanas, G., Leimk{\"u}hler, T., Drettakis, G.: 3d gaussian splatting for real-time radiance field rendering. ACM Transactions on Graphics  \textbf{42}(4) (2023)

\bibitem{khosla2020supervised}
Khosla, P., Teterwak, P., Wang, C., Sarna, A., Tian, Y., Isola, P., Maschinot, A., Liu, C., Krishnan, D.: Supervised contrastive learning. Advances in neural information processing systems  \textbf{33},  18661--18673 (2020)

\bibitem{kollar2022simnet}
Kollar, T., Laskey, M., Stone, K., Thananjeyan, B., Tjersland, M.: Simnet: Enabling robust unknown object manipulation from pure synthetic data via stereo. In: Conference on Robot Learning. pp. 938--948. PMLR (2022)

\bibitem{kuhn1955hungarian}
Kuhn, H.W.: The hungarian method for the assignment problem. Naval research logistics quarterly  \textbf{2}(1-2),  83--97 (1955)

\bibitem{lee2023tta}
Lee, T., Tremblay, J., Blukis, V., Wen, B., Lee, B.U., Shin, I., Birchfield, S., Kweon, I.S., Yoon, K.J.: Tta-cope: Test-time adaptation for category-level object pose estimation. In: Proceedings of the IEEE/CVF Conference on Computer Vision and Pattern Recognition. pp. 21285--21295 (2023)

\bibitem{lin2022category}
Lin, J., Wei, Z., Ding, C., Jia, K.: Category-level 6d object pose and size estimation using self-supervised deep prior deformation networks. In: European Conference on Computer Vision. pp. 19--34. Springer (2022)

\bibitem{lin2017feature}
Lin, T.Y., Doll{\'a}r, P., Girshick, R., He, K., Hariharan, B., Belongie, S.: Feature pyramid networks for object detection. In: Proceedings of the IEEE conference on computer vision and pattern recognition. pp. 2117--2125 (2017)

\bibitem{lin2017focal}
Lin, T.Y., Goyal, P., Girshick, R., He, K., Doll{\'a}r, P.: Focal loss for dense object detection. In: Proceedings of the IEEE international conference on computer vision. pp. 2980--2988 (2017)

\bibitem{liu2023net}
Liu, J., Chen, Y., Ye, X., Qi, X.: Ist-net: Prior-free category-level pose estimation with implicit space transformation. In: Proceedings of the IEEE/CVF International Conference on Computer Vision. pp. 13978--13988 (2023)

\bibitem{liu2022akb}
Liu, L., Xu, W., Fu, H., Qian, S., Yu, Q., Han, Y., Lu, C.: Akb-48: A real-world articulated object knowledge base. In: Proceedings of the IEEE/CVF Conference on Computer Vision and Pattern Recognition. pp. 14809--14818 (2022)

\bibitem{liu2023zero}
Liu, R., Wu, R., Van~Hoorick, B., Tokmakov, P., Zakharov, S., Vondrick, C.: Zero-1-to-3: Zero-shot one image to 3d object. In: Proceedings of the IEEE/CVF International Conference on Computer Vision. pp. 9298--9309 (2023)

\bibitem{liu2020keypose}
Liu, X., Jonschkowski, R., Angelova, A., Konolige, K.: Keypose: Multi-view 3d labeling and keypoint estimation for transparent objects. In: Proceedings of the IEEE/CVF conference on computer vision and pattern recognition. pp. 11602--11610 (2020)

\bibitem{liu2022petr}
Liu, Y., Wang, T., Zhang, X., Sun, J.: Petr: Position embedding transformation for multi-view 3d object detection. In: European Conference on Computer Vision. pp. 531--548. Springer (2022)

\bibitem{liu2022gen6d}
Liu, Y., Wen, Y., Peng, S., Lin, C., Long, X., Komura, T., Wang, W.: Gen6d: Generalizable model-free 6-dof object pose estimation from rgb images. In: European Conference on Computer Vision. pp. 298--315. Springer (2022)

\bibitem{liu2022convnet}
Liu, Z., Mao, H., Wu, C.Y., Feichtenhofer, C., Darrell, T., Xie, S.: A convnet for the 2020s. In: Proceedings of the IEEE/CVF conference on computer vision and pattern recognition. pp. 11976--11986 (2022)

\bibitem{loshchilov2016sgdr}
Loshchilov, I., Hutter, F.: Sgdr: Stochastic gradient descent with warm restarts. arXiv preprint arXiv:1608.03983  (2016)

\bibitem{loshchilov2017decoupled}
Loshchilov, I., Hutter, F.: Decoupled weight decay regularization. arXiv preprint arXiv:1711.05101  (2017)

\bibitem{lunayach2023fsd}
Lunayach, M., Zakharov, S., Chen, D., Ambrus, R., Kira, Z., Irshad, M.Z.: Fsd: Fast self-supervised single rgb-d to categorical 3d objects. arXiv preprint arXiv:2310.12974  (2023)

\bibitem{mees2019self}
Mees, O., Tatarchenko, M., Brox, T., Burgard, W.: Self-supervised 3d shape and viewpoint estimation from single images for robotics. In: 2019 IEEE/RSJ International Conference on Intelligent Robots and Systems (IROS). pp. 6083--6089. IEEE (2019)

\bibitem{mescheder2019occupancy}
Mescheder, L., Oechsle, M., Niemeyer, M., Nowozin, S., Geiger, A.: Occupancy networks: Learning 3d reconstruction in function space. In: Proceedings of the IEEE/CVF conference on computer vision and pattern recognition. pp. 4460--4470 (2019)

\bibitem{mildenhall2021nerf}
Mildenhall, B., Srinivasan, P.P., Tancik, M., Barron, J.T., Ramamoorthi, R., Ng, R.: Nerf: Representing scenes as neural radiance fields for view synthesis. Communications of the ACM  \textbf{65}(1),  99--106 (2021)

\bibitem{mo2019partnet}
Mo, K., Zhu, S., Chang, A.X., Yi, L., Tripathi, S., Guibas, L.J., Su, H.: Partnet: A large-scale benchmark for fine-grained and hierarchical part-level 3d object understanding. In: Proceedings of the IEEE/CVF conference on computer vision and pattern recognition. pp. 909--918 (2019)

\bibitem{park2019deepsdf}
Park, J.J., Florence, P., Straub, J., Newcombe, R., Lovegrove, S.: Deepsdf: Learning continuous signed distance functions for shape representation. In: Proceedings of the IEEE/CVF conference on computer vision and pattern recognition. pp. 165--174 (2019)

\bibitem{shi2021stablepose}
Shi, Y., Huang, J., Xu, X., Zhang, Y., Xu, K.: Stablepose: Learning 6d object poses from geometrically stable patches. In: Proceedings of the IEEE/CVF Conference on Computer Vision and Pattern Recognition. pp. 15222--15231 (2021)

\bibitem{sun2022onepose}
Sun, J., Wang, Z., Zhang, S., He, X., Zhao, H., Zhang, G., Zhou, X.: Onepose: One-shot object pose estimation without cad models. In: Proceedings of the IEEE/CVF Conference on Computer Vision and Pattern Recognition. pp. 6825--6834 (2022)

\bibitem{tian2020shape}
Tian, M., Ang, M.H., Lee, G.H.: Shape prior deformation for categorical 6d object pose and size estimation. In: Computer Vision--ECCV 2020: 16th European Conference, Glasgow, UK, August 23--28, 2020, Proceedings, Part XXI 16. pp. 530--546. Springer (2020)

\bibitem{tochilkin2024triposr}
Tochilkin, D., Pankratz, D., Liu, Z., Huang, Z., Letts, A., Li, Y., Liang, D., Laforte, C., Jampani, V., Cao, Y.P.: Triposr: Fast 3d object reconstruction from a single image. arXiv preprint arXiv:2403.02151  (2024)

\bibitem{wang2021gdr}
Wang, G., Manhardt, F., Tombari, F., Ji, X.: Gdr-net: Geometry-guided direct regression network for monocular 6d object pose estimation. In: Proceedings of the IEEE/CVF Conference on Computer Vision and Pattern Recognition. pp. 16611--16621 (2021)

\bibitem{wang2019normalized}
Wang, H., Sridhar, S., Huang, J., Valentin, J., Song, S., Guibas, L.J.: Normalized object coordinate space for category-level 6d object pose and size estimation. In: Proceedings of the IEEE/CVF Conference on Computer Vision and Pattern Recognition. pp. 2642--2651 (2019)

\bibitem{weng2021captra}
Weng, Y., Wang, H., Zhou, Q., Qin, Y., Duan, Y., Fan, Q., Chen, B., Su, H., Guibas, L.J.: Captra: Category-level pose tracking for rigid and articulated objects from point clouds. In: Proceedings of the IEEE/CVF International Conference on Computer Vision. pp. 13209--13218 (2021)

\bibitem{wu2023omniobject3d}
Wu, T., Zhang, J., Fu, X., Wang, Y., Ren, J., Pan, L., Wu, W., Yang, L., Wang, J., Qian, C., et~al.: Omniobject3d: Large-vocabulary 3d object dataset for realistic perception, reconstruction and generation. In: Proceedings of the IEEE/CVF Conference on Computer Vision and Pattern Recognition. pp. 803--814 (2023)

\bibitem{Xiang_2020_CVPR}
Xiang, F., Qin, Y., Mo, K., Xia, Y., Zhu, H., Liu, F., Liu, M., Jiang, H., Yuan, Y., Wang, H., Yi, L., Chang, A.X., Guibas, L.J., Su, H.: Sapien: A simulated part-based interactive environment. In: Proceedings of the IEEE/CVF Conference on Computer Vision and Pattern Recognition (CVPR) (June 2020)

\bibitem{xie2021toward}
Xie, H., Yao, H., Zhou, S., Zhang, S., Tong, X., Sun, W.: Toward 3d object reconstruction from stereo images. Neurocomputing  \textbf{463},  444--453 (2021)

\bibitem{xie2022neural}
Xie, Y., Takikawa, T., Saito, S., Litany, O., Yan, S., Khan, N., Tombari, F., Tompkin, J., Sitzmann, V., Sridhar, S.: Neural fields in visual computing and beyond. In: Computer Graphics Forum. vol.~41, pp. 641--676. Wiley Online Library (2022)

\bibitem{ze2022category}
Ze, Y., Wang, X.: Category-level 6d object pose estimation in the wild: A semi-supervised learning approach and a new dataset. Advances in Neural Information Processing Systems  \textbf{35},  27469--27483 (2022)

\bibitem{zhang2022transnet}
Zhang, H., Opipari, A., Chen, X., Zhu, J., Yu, Z., Jenkins, O.C.: Transnet: Category-level transparent object pose estimation. In: European Conference on Computer Vision. pp. 148--164. Springer (2022)

\bibitem{zhang2024generative}
Zhang, J., Wu, M., Dong, H.: Generative category-level object pose estimation via diffusion models. Advances in Neural Information Processing Systems  \textbf{36} (2024)

\bibitem{zhang2022self}
Zhang, K., Fu, Y., Borse, S., Cai, H., Porikli, F., Wang, X.: Self-supervised geometric correspondence for category-level 6d object pose estimation in the wild. arXiv preprint arXiv:2210.07199  (2022)

\end{thebibliography}
\end{document}